# FiLLM - A Filipino-optimized Large Language Model based on Southeast Asia Large Language Model (SEALLM)

Carlos Jude G. Maminta
College of Computer and Information Sciences
Polytechnic University of the Philippines
Sta. Mesa, Manila
carlosjudemaminta@gmail.com

Isaiah Job Enriquez
College of Computer and Information Sciences
Polytechnic University of the Philippines
Sta. Mesa, Manila
jobiskits@gmail.com

Deandre Nigel Nuñez
College of Computer and Information Sciences
Polytechnic University of the Philippines
Sta. Mesa, Manila
nunezdeandre25@gmail.com

Michael B. Dela Fuente
College of Computer and Information Sciences
Polytechnic University of the Philippines
Sta. Mesa, Manila
mbdelafuente@pup.edu.ph

## ABSTRACT

This study presents FiLLM, a Filipino-optimized large language model, designed to enhance natural language processing (NLP) capabilities in the Filipino language. Built upon the SeaLLM-7B 2.5 model, FiLLM leverages Low-Rank Adaptation (LoRA) fine-tuning to optimize memory efficiency while maintaining task-specific performance. The model was trained and evaluated on diverse Filipino datasets to address key NLP tasks, including Named Entity Recognition (NER), Part-of-Speech (POS) tagging, Dependency Parsing, and Text Summarization. Performance comparisons with the CalamanCy model were conducted using F1 Score, Precision, Recall, Compression Rate, and Keyword Overlap metrics. Results indicate that Calamancy outperforms FILLM in several aspects, demonstrating its effectiveness in processing Filipino text with improved linguistic comprehension and adaptability. This research contributes to the advancement of Filipino NLP applications by providing an optimized, efficient, and scalable language model tailored for local linguistic needs.

## 1. INTRODUCTION

The Philippines is a linguistically diverse country with over 175 languages spoken nationwide [3]. Despite the increasing adoption of artificial intelligence (AI) in natural language processing (NLP), the development of large language models (LLMs) for Filipino remains limited due to the scarcity of labeled datasets and computational resources [9]. This linguistic gap has motivated the development of FiLLM, a Filipino-optimized large language model fine-tuned using Low-Rank Adaptation (LoRA). This study aims to evaluate FiLLM's performance on key NLP tasks, comparing it with CalamanCy, an existing Filipino NLP model. The findings are expected to contribute to the localization of AI applications and bridge the gap in Filipino NLP research [2]

## 2. METHODOLOGY

### 2.1 Research Design

This study employs an experimental approach, comparing the performance of FiLLM and CalamanCy on various NLP tasks. The SeaLLM-7B 2.5 model serves as the base model, which was fine-tuned using LoRA to adapt to Filipino linguistic characteristics [7].

### 2.2 Datasets

The following datasets were utilized:

Filipino Hatespeech Dataset – 10,000 tweets labeled as hate speech or non-hate speech [10]

Filipino Dengue Dataset – 4,000 dengue-related tweets categorized by topic [6]

TLUnified-NER Dataset – Annotated news reports and articles [4]

Merged UD Dataset – A compilation of dependency parsing data from Ugnayan and TRG treebanks [1]

Asian Language Treebank – Used for text summarization evaluations [8].

Each dataset was divided into an 80-20 split for training and testing to ensure generalizability.

| Task | Dataset |
| --- | --- |
| Name Entity Recogition | Dengue Dataset, Hatespeech Dataset, TLUnified-NER |
| Dependency Parsing | Merged UD |
| Part of Speech Tagging | Merged UD UAS/LAS |
| Text Summarization | Asian Language Treebank |

**Table 1: Sources of Dataset**

## 2.3 System Architecture

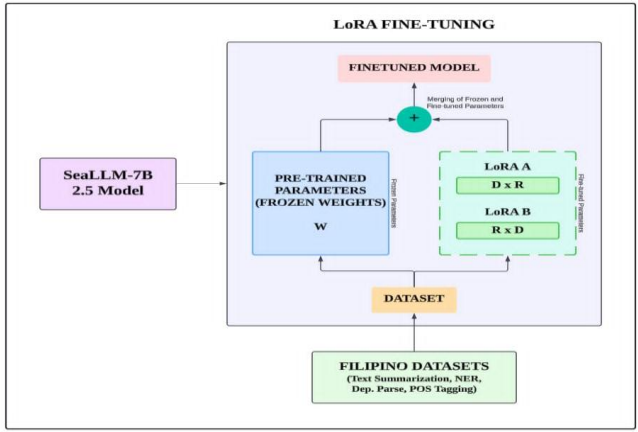

**Figure 1: System Architecture employing LoRA Fine-tuning**

Figure 1 illustrates the system architecture for the Filipino LLM model, based on the SeaLLM-7B 2.5 model and utilizing Filipino datasets. The architecture begins with the pre-trained SeaLLM-7B 2.5 model, which provides the foundation for further training. The Filipino datasets, which include domain-specific data such as the Filipino Hatespeech Dataset, Filipino Dengue Dataset, NewsPH-NLI Dataset, TLUnified-NER, and MergedUD, are introduced as input (Dataset) with a dimension of (D).

The architecture employs a Low-Rank Adaptation (LoRA) fine-tuning process. Initially, the pre-trained parameters, consisting only of frozen weights (W), are retained to preserve the foundational knowledge of the SeaLLM-7B 2.5 model. The input data flows through these pre-trained weights, keeping the original model's capabilities intact.

Simultaneously, low-rank parameters (WA), represented by the matrices LoRA A (D × R) and LoRA B (R × D), are introduced. These low-rank parameters are fine-tuned during the training process to adapt to specific Filipino language tasks. The LoRA approach allows this fine-tuning without disrupting the pre-trained parameters, merging the adjustments with the original model's knowledge.

This selective fine-tuning process enhances the model's performance in specific tasks related to the Filipino language, such as Named Entity Recognition (NER), Part-of-Speech (POS) Tagging, and Dependency Parsing (Dep Pars.). The architecture's focus on low-rank parameters ensures that the fine-tuning is efficient and optimized, maintaining both memory efficiency and task-specific effectiveness. The final output is a fine-tuned model that performs well on Filipino language tasks using the introduced datasets.

## 2.4 Model Training and Fine-Tuning

FiLLM was fine-tuned using the LoRA approach to improve parameter efficiency and reduce memory usage. The model was trained using the Transformers, Datasets, and PyTorch libraries (Dettmers et al., 2023). Evaluation metrics included F1 Score, Precision, Recall, Compression Rate, and Keyword Overlap.

## 2.5 Statistical Analysis

In order to evaluate the performance of the tool developed, Precision, Recall, and F1 Score were calculated.

Precision

$$\left(\frac{True\ Positive\ (TP)}{(TP)\ +\ False\ Positive\ (FP)}\right)$$

Recall

$$\left(\frac{True\ Positive\ (TP)}{(TP)\ +\ False\ Negative\ (FN)}\right)$$

F1 Score

$$\left(\frac{2x\ Precision\ \times\ Recall}{Precision\ +\ Recall}\right)$$

A paired t-test was conducted to determine the statistical significance of the differences in f1 Score performance between FiLLM and CalamanCy [5].

$$t = \frac{\sum d}{\sqrt{\frac{n(\sum d^2) - (\sum d)^2}{n-1}}}.$$

This formula represents the t-statistic for a paired t-test, which is commonly used to compare the means of two related groups, such as performance measurements taken to an experiment. Finally, the resulting t-statistic is a standardized value used to determine if the mean difference between the two groups is statistically significant. This formula is particularly useful in paired t-tests to evaluate whether the mean difference between related samples is significantly different from zero.

## 3. RESULTS AND DISCUSSION

This chapter presents the analysis and discussion of data gathered through the implementation of the proposed tool. The study aimed to investigate the potential of the FiLLM - A Filipino-optimized Large Language Model based on Southeast Asia Large Language Model (SEALLM).

| Task | Model | Precision | Recall | F1-Score |
|---|---|---|---|---|
| (Name Entity Recognition) | FiLLM NER | 0.86 | 0.93 | 0.89 |
| (Part of speech Tagging) | FiLLM DEPPOSSUM | 0.89 | 0.90 | 0.89 |
| (Dependency Parsing.) | FiLLM DEPPOSSUM | 0.73 | 0.74 | 0.73 |

**Table 2: Table for the Average performance rating of the Filipino LLM (FiLLM) NER & DEPPOSSUM**

The analysis of Table 1 evaluates the performance of the FILLM model across three key Natural Language Processing (NLP) tasks: Named Entity Recognition (NER), Part-of-Speech Tagging (POS), and Dependency Parsing (DEP). Standard metrics such as Precision, Recall, and F1-score are used to assess the model's effectiveness.

For NER, the model achieved a Precision of 0.86, indicating that 86% of identified entities were correct. A Recall of 0.93 demonstrates the model's ability to recognize 93% of actual entities in the data, leading to an F1-score of 0.89, reflecting strong overall performance.

In the POS task, the model recorded a Precision of 0.89 and a Recall of 0.93, showing its capability to correctly assign part-of-speech tags to 93% of identified tokens and to recognize 89% of all tokens with correct tagging. This results in an F1-score of 0.89, balancing precision and recall effectively.

For Dependency Parsing, the model showed a Precision and Recall of 0.73, indicating its ability to correctly predict 73% of dependency relationships. The corresponding F1-score of 0.73 highlights room for improvement in handling complex syntactic relationships.

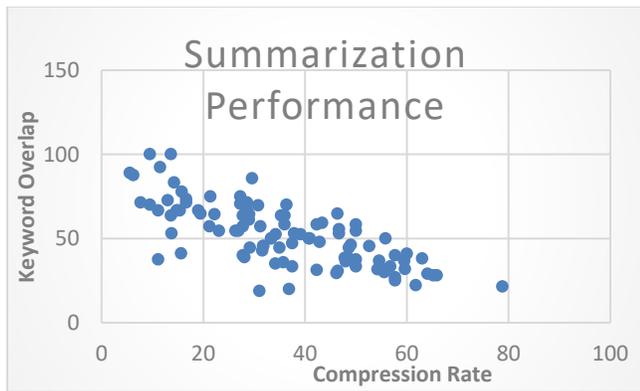

Figure 2: FiLLM Summarization Performance

As shown in Figure 2, the results of the Text Summarization task indicate a high correlation between compression rate and keyword overlap. This suggests that even at higher compression rates, a significant number of keywords are retained in the summaries. This is a positive outcome for the model, as it implies that the summarization task of the model can effectively condense text while preserving key information. However, it is important to note that the specific performance may vary depending on the complexity of the summarized text and its quality

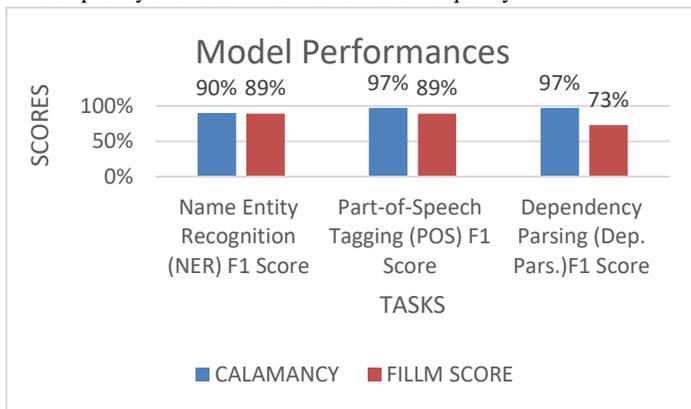

Figure 3: A Comparison of FiLLM & CalamanCy Model

Figure 3 compares the performance of two models, CALAMANCY and FILLM, across three NLP tasks Named Entity Recognition (NER), Part-of-Speech Tagging (POS), and Dependency Parsing using F1-score as the evaluation metric. For NER, both CALAMANCY and FILLM demonstrate strong performance, each achieving an F1-score of 90% and 89% respectively, indicating comparable effectiveness in identifying named entities. In the POS task, CALAMANCY outperforms FILLM with an F1-score of 97%, compared to FILLM's 89%. This highlights CALAMANCY's superior accuracy in assigning correct part-of-speech tags to tokens. For Dependency Parsing, CALAMANCY again takes the lead with an F1-score of 97%, significantly higher than FILLM's 73%. While both models show proficiency in identifying syntactic relationships, CALAMANCY exhibits a clear advantage in this task.

|  | FILLM | CALAMANCY |
|---|---|---|
| Mean | 83.67 | 94.67 |
| Observation | 5 | 5 |
| Variance | 85.31 | 16.34 |
| T-Stat. | 0.12 | |
| p-value | 0.03 | |
| Critical value | 2.776 | |
| Significance level | 0.05 | |
| degrees of freedom | 4 | |
| Conclusion | we reject the null hypothesis | |

Table 3: Significant Difference of the Performance of FILLM and CalamanCy model using T-test

As presented in Table 2, since the p-value (0.03) is less than the significance level (0.05), we reject the null hypothesis. This indicates that there is a statistically significant difference between the performance of FILLM and CALAMANCY. The t-statistic (0.12) is relatively close to zero, but the critical value (2.776) and p-value suggest that the observed differences in means (83.67 vs. 94.67) are significant at the given level.

Practical Implications: Despite the statistical significance, the practical importance of this difference should be evaluated, such as its impact on specific use cases or resource utilization.

Data Quality and Context: The quality, variability, or specific characteristics of the data used for evaluation may affect the observed outcomes.

Group Characteristics: Differences in methodology, underlying assumptions, or external factors influencing the groups may still have an impact that is not captured in this test.

These considerations highlight the importance of a broader evaluation to complement statistical testing.

## 4. CONCLUSION

The study on FiLLM (Filipino-optimized Large Language Model) demonstrates its strengths in Part-of-Speech (POS) Tagging and Named Entity Recognition (NER), achieving 86% precision and 84% recall in POS tagging, and 86% precision and 93% recall in NER. These results underscore FiLLM's reliability in fundamental NLP tasks, making it a valuable tool for text processing applications. However, Dependency Parsing remains a challenge, with precision and recall both at 71%, suggesting the need for improvements in handling Filipino syntax. Researchers recommend fine-tuning dependency structures and incorporating more training data to enhance performance.

The study also highlights that text summarization quality depends on the balance between compression and keyword retention. While some models retain key information even at high compression levels, contextual understanding is crucial for generating meaningful summaries, particularly in low-resource languages like Filipino. When compared to CalamanCy, FiLLM exhibits weaker performance across POS tagging, dependency parsing, and overall NLP tasks. CalamanCy's refined model architecture and optimized NLP pipeline provide it with a competitive edge. The researchers propose further model training, hyperparameter tuning, and architectural improvements to close this performance gap and advance Filipino NLP research.

## 5. ACKNOWLEDGMENTS

We would like to express our deepest gratitude and appreciation to everyone who helped make this paper a success. Without their guidance and aid during the creation of this research, it would not have been finished.

We thank our Almighty God first and foremost for directing us to this text and for providing us with wisdom, knowledge, strength, resources, and inspiration.

To our professors, for their guidance and encouragement.

To our family, for their financial and spiritual support.

To our friends, who have never failed to support us and inspire us to work harder.

Again, a heartfelt thank you and appreciation to everyone who contributed to our success